\documentclass{proc}

\usepackage{times}
\usepackage{graphicx}
\usepackage[hang,scriptsize]{subfigure}
\usepackage{latexsym}

\makeatletter
\def\@maketitle{%
  \vbox to 2.3in{%
    \hsize\textwidth
    \linewidth\hsize
    \vspace*{1.5cm}
    \centering
    {\bfseries\huge \@title \par}
    \vskip 2em
    {\large \begin{tabular}[t]{c}\@author \end{tabular}\par}
    \vfill}    \vspace*{1.0cm}
}
\renewcommand\section{\@startsection {section}{1}{\z@}%
     {.7\baselineskip plus\baselineskip}{.5\baselineskip}
                                   {\normalfont\Large\bfseries}}
\renewcommand\section{\@startsection {section}{1}{\z@}%
      {.5\baselineskip\@plus.7\baselineskip}{.3\baselineskip}%
                                   {\normalfont\Large\bfseries}}
\renewcommand\subsection{\@startsection{subsection}{2}{\z@}%
       {.5\baselineskip\@plus.7\baselineskip}{.3\baselineskip}%
                                   {\normalfont\large\bfseries}}
\renewcommand\subsubsection{\@startsection{subsubsection}{3}{\z@}%
      {.5\baselineskip\@plus.7\baselineskip}{.3\baselineskip}%
                                     {\normalfont\normalsize\bfseries}}
\makeatother

\renewenvironment{abstract}%
  {\normalfont
    \list{}{\labelwidth0pt
      \leftmargin0pt \rightmargin\leftmargin
      \listparindent\parindent \itemindent0pt
      \parsep0pt
      
    }%
    \item[\hskip\labelsep\bfseries\abstractname\enspace --] \itshape%
}{%
  \endlist}

\newcommand{\keywordsname}{Keywords}
\newenvironment{keywords}%
  {\normalfont
    \list{}{\labelwidth0pt
      \leftmargin0pt \rightmargin\leftmargin
      \listparindent\parindent \itemindent0pt
      \parsep0pt
      }%
    \item[\hskip\labelsep\bfseries\keywordsname:]}{\endlist}

\def\bsigma{\mbox{\boldmath $\sigma$}}

\def\bxi{\mbox{\boldmath $\xi$}}

\def\V{{\bf V}}
\def\W{{\bf W}}
\def\X{{\bf X}}

\def\Z{{\bf Z}}

\def\p{{\bf p}}

\def\x{{\bf x}}

\def\z{{\bf z}}

\def\0{{\bf 0}}

\begin{document}

\title{Multi-Target Particle Filtering for the Probability Hypothesis Density\vspace{-0.4cm}}
\author{Hedvig Sidenbladh\\[0.2cm]
Department of Data and Information Fusion\\
Division of Command and Control Systems\\
Swedish Defence Research Agency\\
SE--172 90 Stockholm, Sweden\\
{\tt hedvig@foi.se}}
\date{}
\maketitle

\begin{abstract}

When tracking a large number of targets, it is often computationally expensive 
to represent the full joint distribution over target states. 
In cases where the targets move independently, each target can instead be 
tracked with a separate filter. However, this leads to a model-data 
association problem.
Another approach to solve the problem with computational complexity is to 
track only the first moment of the joint distribution, 
the {\em probability hypothesis density (PHD)}. The integral of this 
distribution over any area $S$ is the expected number of targets within $S$. 
Since no record of object identity is kept, the model-data association problem 
is avoided.

The contribution of this paper is a particle filter implementation of the PHD 
filter mentioned above. This PHD particle filter is applied to tracking of 
multiple vehicles 
in terrain, a non-linear tracking problem. Experiments show that the filter
can track a changing number of vehicles robustly, achieving near-real-time 
performance.
\end{abstract}


\section{Introduction}
\label{sec:introduction}

When tracking multiple targets in general, the size of the state-space for the 
joint distribution over target states grows exponentially with the number of 
targets. When the number of targets is large, this makes it impossible in 
practice to maintain the joint distribution over target states.
However, if the targets can be assumed to move independently, the joint 
distribution does not have to be maintained. A straight-forward method is to 
assign a separate filter to each target \cite{fortmann83,reid79}. A drawback 
with this approach is that it leads to a model-data association problem 
\cite{fortmann83}.

A mathematically principled alternative to the separate filter approach is to 
propagate only 
the first moment of the joint distribution, the {\em probability hypothesis 
density (PHD)} \cite{mahlerSPIE02,mahler-zajic01}. This entity is described in
Section \ref{sec:phd-phd}, and is defined over the state-space for one target. 
It has the property that for each sub-area $S$ in the state-space, the 
integral of the PHD over $S$ is the expected number of targets within this 
area. Thus, peaks in the PHD can be regarded as estimated target states.
Since the identities of objects are not maintained, there is no model-data 
association problem.

The main contribution of this paper is a particle filter 
\cite{gordon93,isard-blakeIJCV98} implementation of 
PHD tracking, the {\em PHD particle filter}. The PHD particle filter 
implementation is described in Section \ref{sec:phd-particle}.

Particle filtering (Section \ref{sec:particle-particle}) is suited 
for tracking with non-linear and non-Gaussian motion models. 
Here, the PHD particle filter 
is applied to tracking of multiple vehicles in terrain (Section 
\ref{sec:application}), a problem which is
highly non-linear due to the terrain (Section 
\ref{sec:application-motion}). The vehicles are observed by humans 
situated in the terrain. Two things should be noted about this application: 
Since the observations originate from humans rather 
than automatic sensors, the degree of missing observations is much higher than 
the degree of spurious observations. Furthermore, the time-scale is quite long 
-- one time-step is on the order of a few seconds. Thus, the relatively high 
computational complexity of particle filters compared to, e.g., Kalman filters 
provides less of a problem for real-time implementation than it would in many 
other applications.
Experiments in Section \ref{sec:experiments} show the PHD particle filter to 
be a fast, efficient and robust alternative to tracking of the full joint 
distribution over targets.

\section{Related work}
\label{sec:related}

\paragraph{Multi-target tracking.} 
The problem of tracking multiple targets is more difficult than the tracking 
of a single target in two aspects. 

If the number of targets is known and constant over time, the problem of 
tracking 
multiple targets is just a natural extension of single target tracking in the
state-space spanned by all object state-spaces.  
However, if the number of targets is unknown or varies over time, the number 
of targets, $N$, is itself a (discrete) random variable, and a part of the 
state-space. Since the dimensionality of the state-space varies with $N$ 
(e.g., two targets are described by twice as many parameters as a single 
target), it is not possible to compare two states of different value $n$ of 
$N$ using ordinary Bayesian statistics. One way to address this problem 
\cite{fortmann83,hue02} is to 
estimate $N$ separately from the rest of the state-space, and then, given 
this, estimate the other state variables knowing the size of the state-space. 
Another \cite{stone02} is to assume 
$N$ known and constant, and model some of the targets as ``hidden''. A third 
approach \cite{ballantyne01,isard-maccormick01} is to do the likelihood 
evaluation in a space of constant dimensionality (the image space), thus 
avoiding the problem of comparing spaces of different dimensionality.
However, the problem can also be addressed by employing {\em finite set 
statistics (FISST)} \cite{goodman97,mahler-mono} which is an extension of Bayesian 
analysis to incorporate comparisons of between state-spaces of different 
dimensionality.
Thus, a distribution over $N$ can be estimated with the rest of the 
state-space. FISST has been used extensively for tracking 
\cite{mahler-mono,mahlerSPIE02,mahler-zajic01,musick98},
mainly implemented as a set of Kalman or $\alpha$-$\beta$-$\gamma$-filters.
The particle filter presented here is formulated within this framework.

The second problem with multi-target 
tracking in general is that the size of 
the state-space grows exponentially with the number of targets. Even with 
tracking algorithms that very efficiently search the state-space, it is not 
possible to estimate the joint distribution over a large number of targets 
with a limited computational effort. 
However, if the targets move independently, simplifications can be introduced.
One approach is simply to track each target using a separate filter, 
e.g.~\cite{fortmann83,reid79}. This simplification allows for tracking of a 
large number of targets, but leads to a model-data association problem, 
addressed by e.g.~joint probabilistic data association (JPDA) 
\cite{fortmann83}. 
To avoid this problem, Mahler and Zajic \cite{mahlerSPIE02,mahler-zajic01} 
formulate an algorithm for propagating a combined density (PHD) over all 
targets, instead of modeling the probability density function (pdf) for 
each individual target. We present a particle filter implementation of this
PHD filter.

\vspace{-0.1cm}
\paragraph{Terrain tracking.} 
The problem of tracking in terrain differs from, e.g., air target tracking in 
that it is non-linear and non-Gaussian, due to the variability in the terrain.
This makes linear Kalman tracking approaches like Interacting Multiple Models 
(IMM) \cite{mazor98} inappropriate, since it is difficult to model the terrain 
influence in a general manner. However, in a simplified environment, such as a 
terrain map with only on/off road information, IMM-based approaches are 
successful \cite{ke00}. Another type of approach is to formulate the 
terrain as a potential field \cite{kastella00,sodtke-llinas01} or 
an HMM \cite{ke00}, where the transition probabilities correspond to 
terrain movability in that area. This allows for modeling of the 
non-linearities in the terrain. However, the potential field approach is 
computationally expensive \cite{kastella00}. Furthermore, a comparison 
\cite{ke00} between the HMM and an IMM filter shows the IMM 
approach to be more efficient in a linearized situation.

We take a slightly different approach. To cope with the non-linearities of 
the terrain tracking problem in a mathematically principled way, we use 
particle filtering (also known as bootstrap filtering \cite{gordon93} or 
Condensation \cite{isard-blakeIJCV98}), which has proven useful 
for tracking with non-linear and non-Gaussian models of motion and 
observations.

\section{Bayesian filtering}
\label{sec:particle}

We start by describing the formulation of the discrete-time tracking problem 
for a single target, with exactly one observation in each time-step.

In a Bayesian filter, the tracking problem is formulated as an iterative 
implementation of Bayes' theorem. All information about the state of the 
tracked target can be deduced from the {\em posterior distribution} 
$f_{\X_t\,|\,\Z_{1:t}}(\x_t\,|\,\z_{1:t})$ over states $\X_t$, conditioned 
on the history of observations $\Z_{1:t}$ from time $1$ up to time $t$. 
The filter consists of two steps, prediction and observation:

\vspace{-0.1cm}
\paragraph{Prediction.} In the prediction step, the {\em prior distribution} 
$f_{\X_t\,|\,\Z_{1:t-1}}(\x_t\,|\,\z_{1:t-1})$ at time $t$ is deduced from 
the posterior at time $t-1$ as
\begin{eqnarray}
\label{eq:particle-prior}
f_{\X_t\,|\,\Z_{1:t-1}}(\x_t\,|\,\z_{1:t-1}) = \hspace{3.85cm}\nonumber\\ 
  \int f_{\X_t\,|\,\X_{t-1},\Z_{1:t-1}}(\x_t\,|\,\x_{t-1},\z_{1:t-1})
\hspace{1.0cm}\nonumber\\
       ~f_{\X_{t-1}\,|\,\Z_{1:t-1}}(\x_{t-1}\,|\,\z_{1:t-1})~d\x_{t-1} 
\end{eqnarray}
where the probability density function (pdf) 
$f_{\X_t\,|\,\X_{t-1},\Z_{1:t-1}}(\x_t\,|\,\x_{t-1},\z_{1:t-1})$ is defined by
a model of motion in its most general form. 

Often, however, the state at time $t$ is generated from the previous state 
according to the model
\begin{equation}
\label{eq:particle-phi}
\X_t = \phi(\X_{t-1}, \W_t)
\end{equation}
where $\W_t$ is a noise term independent of $\X_{t-1}$. This gives 
$f_{\X_t\,|\,\X_{t-1},\Z_{1:t-1}}(\x_t\,|\,\x_{t-1},\z_{1:t-1}) \equiv 
f_{\X_t\,|\,\X_{t-1}}(\x_t\,|\,\x_{t-1})$, with no dependence on the 
history of observations $\z_{1:t-1}$.

\vspace{-0.1cm}
\paragraph{Observation.} In each time-step, observations of the state are 
assumed generated from the model
\begin{equation}
\label{eq:particle-h}
\Z_t = h(\X_t, \V_t)
\end{equation}
where $\V_t$ is a noise term independent of $\X_t$. From this model, the 
{\em likelihood} 
$f_{\Z_t\,|\,\X_t}(\z_t\,|\,\x_t)$ is derived. The posterior at time $t$ is 
computed from the prior (Eq (\ref{eq:particle-prior})) and the likelihood 
according to Bayes' rule:
\begin{eqnarray}
\label{eq:particle-bayes}
f_{\X_t\,|\,\Z_{1:t}}(\x_t\,|\,\z_{1:t}) \propto \hspace{4.4cm}\nonumber\\ 
  f_{\Z_t\,|\,\X_t}(\z_t\,|\,\x_t) 
  ~f_{\X_{t}\,|\,\Z_{1:t-1}}(\x_{t}\,|\,\z_{1:t-1}) ~.
\end{eqnarray}
To conclude, the posterior pdf at time $t$ is calculated from the previous 
posterior at $t-1$, the motion model, and the observations at time $t$ 
according to 
Eqs (\ref{eq:particle-prior}) and (\ref{eq:particle-bayes}).
The iterative filter formulation requires a known initial 
posterior pdf $f_{\X_0\,|\,\Z_{0}}(\x_0\,|\,\z_{0}) \equiv f_{\X_0}(\x_0)$.

\subsection{Particle implementation}
\label{sec:particle-particle}

If the shape of the posterior distribution is close to Gaussian, and the 
functions $h(.)$ and $\phi(.)$ linear, the system can be 
modeled analytically in an efficient manner, e.g.~as a Kalman filter.
However, for non-linear models of motion and observation, the posterior 
distribution will have a more complex shape, often with several maxima. In 
these cases, a Kalman filter is no longer applicable.

Particle filtering, also known as bootstrap filtering 
\cite{gordon93} or Condensation \cite{isard-blakeIJCV98}, 
has proven to be a useful tool for Bayesian tracking with non-linear models of 
motion and observation. Particle filtering is a 
sequential Monte Carlo method. For an overview of the 
state of the art in applications of particle filters, see \cite{doucet01}.

The posterior is represented by a set of $\mathcal{N}$ state hypotheses, or 
particles $\{\bxi_t^1,\ldots,\bxi_t^\mathcal{N}\}$. The density of particles 
in a certain point in state-space represents the posterior density in that 
point \cite{gordon93,isard-blakeIJCV98}. A time-step proceeds as follows:

\vspace{-0.1cm}
\paragraph{Prediction.} The particles 
$\{\bxi_{t-1}^1,\ldots ,\bxi_{t-1}^\mathcal{N}\}$, 
representing 
$f_{\X_{t-1}\,|\,\Z_{1:t-1}}(\x_{t-1}\,|\,\z_{1:t-1})$, are propagated 
in time by sampling from the dynamical model 
$f_{\X_t\,|\,\X_{t-1}}(\x_t\,|\,\bxi^s_{t-1})$ for $s = 1,\ldots,\mathcal{N}$. 
The propagated particles, 
$\{\tilde{\bxi}_t^1,\ldots ,\tilde{\bxi}_t^\mathcal{N}\}$, 
represent the prior 
$f_{\X_t\,|\,\Z_{1:t-1}}(\x_t\,|\,\z_{1:t-1})$ at time $t$. 

\vspace{-0.1cm}
\paragraph{Observation.} Given the new observation $\z_t$ of $\Z_t$, each 
propagated particle $\tilde{\bxi}_t^s$ is assigned a 
weight $\pi_t^s \propto f_{\Z_t\,|\,\X_t}(\z_t\,|\,\tilde{\bxi}^s_t)$. The 
weights are thereafter normalized to sum to one. 

\vspace{-0.1cm}
\paragraph{Resampling.} Now, $\mathcal{N}$ new particles are sampled 
from the set of particles with attached weights, 
$\{(\tilde{\bxi}_t^1,\pi_t^1),\ldots ,
   (\tilde{\bxi}_t^\mathcal{N},\pi_t^\mathcal{N})\}$. The frequency 
with which each particle is resampled is proportional to the weight 
(Monte Carlo sampling). 
The result is a particle set with equal weights, 
$\{\bxi_t^1,\ldots ,\bxi_t^\mathcal{N}\}$,
representing the posterior distribution at time $t$.

\section{FISST multi-target filtering}
\label{sec:phd}
 
We now extend the single-target particle filter to comprise an unknown and 
varying number of targets. The set of tracked objects at time $t$ is a random 
set \cite{goodman97,mahler-mono} $\Gamma_t = \{\X_t^1, \ldots ,\X_t^{N^X_t}\}$, where $\X_t^i$ is the state 
vector of object $i$ and $N^X_t$ is the number of objects in the set. 
A certain outcome of the random set 
$\Gamma_t$ is denoted $X_t = \{\x_t^1, \ldots ,\x_t^{n^X_t}\}$.
Similarly, the set of observations received at time $t$ is a random set 
$\Sigma_t = \{\Z_t^1, \ldots ,\Z_t^{N^Z_t}\}$, where $N^Z_t$ can be larger
than, the same as, or smaller than $N^X_t$. A certain outcome of the random 
set $\Sigma_t$ is denoted $Z_t = \{\z_t^1, \ldots ,\z_t^{n^Z_t}\}$.

Using these random set representations, the multi-target version of Eq 
(\ref{eq:particle-bayes}) is 
\cite{goodman97,mahler-mono}
\begin{eqnarray}
\label{eq:phd-fullbayes}
f_{\Gamma_t\,|\,\Sigma_{1:t}}(X_t\,|\,Z_{1:t}) \propto \hspace{4.4cm}\nonumber\\ 
  f_{\Sigma_t\,|\,\Gamma_t}(Z_t\,|\,X_t) 
  ~f_{\Gamma_{t}\,|\,\Sigma_{1:t-1}}(X_{t}\,|\,Z_{1:t-1})
\end{eqnarray}
where $f_{\Gamma_t\,|\,\Sigma_{1:t}}(X_t\,|\,Z_{1:t})$ is a multi-target 
posterior belief density function, $f_{\Sigma_t\,|\,\Gamma_t}(Z_t\,|\,X_t)$ 
multi-target, multi-observation likelihood, and 
$f_{\Gamma_{t}\,|\,\Sigma_{1:t-1}}(X_{t}\,|\,Z_{1:t-1})$ a multi-target prior.
These densities are defined using finite set statistics (FISST). Details on 
FISST can be found in \cite{goodman97},
while a general particle formulation is presented in 
\cite{sidenbladhWOMOT03,sidenbladhTSP03}.

\subsection{PHD filtering}
\label{sec:phd-phd}

For a large number of targets, the computational complexity of 
Eq (\ref{eq:phd-fullbayes}) will be very high due to the size of the state-space
(see also discussion in Section \ref{sec:related}). However, if the signal to 
noise ratio (SNR) is high and 
the targets move independently of each other, the full posterior
$f_{\Gamma_t\,|\,\Sigma_{1:t}}(X_t\,|\,Z_{1:t})$ can in each time step be 
approximately recovered from the first moment of this distribution, the  
probability hypothesis density (PHD) \cite{mahler-zajic01}:
\begin{equation}
\label{eq:phd-phd}
D_{\X_t\,|\,\Sigma_{1:t}}(\x_t\,|\,Z_{1:t}) = 
\int f_{\Gamma_t\,|\,\Sigma_{1:t}}(\{\x_t\} \cup Y\,|\,Z_{1:t})~\delta Y 
\end{equation}
which is defined over the state-space $\Theta$ of one target, instead of the 
much larger space $\Theta^{N_t^X}$ in which the full posterior 
$f_{\Gamma_t\,|\,\Sigma_{1:t}}(X_t\,|\,Z_{1:t})$ live. This means that the 
computational cost of propagating the PHD over time is much lower than 
propagating the full posterior.

The PHD has the properties that, for any subset $S \subseteq \Theta$, the 
integral of the PHD over $S$ is the expected number of objects in $S$ at time 
$t$:
\begin{equation}
\label{eq:phd-expected}
E[|\Gamma_t \cap S|] = 
\int_S D_{\X_t\,|\,\Sigma_{1:t}}(\x_t\,|\,Z_{1:t})~d\x_t ~.
\end{equation}
In other words, it will have local maxima approximately at the locations of 
the targets. The integral of the PHD over $\Theta$ is the expected number of 
targets, $n_t^X$. 

To find the target locations, a mixture of Gaussians is fitted to the PHD in 
each time step. A local maximum is then found as the mean of a Gaussian in the 
mixture.

We now describe one time-step in the PHD filter. The PHD can not be exactly 
maintained over time \cite{mahler-zajic01}; how good the approximative 
estimation of the PHD is 
depends on the SNR. In the description below, $\hat{D}$ is used to denote an 
approximately estimated PHD \cite{mahler-zajic01}.

\vspace{-0.1cm}
\paragraph{Prediction.} The temporal model of the targets include birth 
(appearance of a target in the field of view), 
death (disappearance of a target from the field of view) and temporal 
propagation. Probability of target death is $p_D$ and of target 
birth $p_B$. Both these probabilities are state independent. 

Target hypotheses are, as in the single target case, propagated from earlier 
hypotheses according to the dynamical model in Eq (\ref{eq:particle-phi}), 
which defines the motion 
pdf $f_{\X_t\,|\,\X_{t-1}}(\x_t\,|\,\x_{t-1})$, a special case of the general 
motion pdf in Eq (\ref{eq:particle-prior}).

In \cite{mahlerSPIE02}, target hypotheses are assumed to be born from a 
uniform distribution over $\Theta$. Here, to better explore the state-space, 
target hypotheses are born from  observations at the previous time instant. 
This is possible if the observation function $h(.)$ 
(Eq (\ref{eq:particle-h})) can be inverted with respect to $\X_t$:\footnote{In 
general, $h^{-1}_{\X_t}(.)$ 
exists for sensors for which the observation space $\Theta_o$ is the same as 
the state space $\Theta$. Negative examples, for which $h^{-1}_{\X_t}(.)$ is 
often impossible to obtain, are image sensors.}
\begin{equation}
\label{eq:phd-birth}
\X_t = \phi(h_{\X_t}^{-1}(\Z_{t-1}, \V_{t-1}), \W_t)~.
\end{equation}
This model defines the birth pdf $f_{\X_t\,|\,\Z_{t-1}}(\x_t\,|\,\z_{t-1})$
which also is a special case of the motion pdf in Eq (\ref{eq:particle-prior}).

In the multi-target case, there is a random set of observations
$\Sigma_t = \{\Z_t^1, \ldots ,\Z_t^{N^Z_t}\}$. To take all observations into 
account for target birth, a birth PHD is defined from the
set of birth pdf:s as 
\begin{equation}
\label{eq:phd-obs}
D_{\X_t\,|\,\Sigma_{t-1}}(\x_t\,|\,Z_{t-1}) = 
\sum_{\z^i_{t-1} \in Z_{t-1}} f_{\X_t\,|\,\Z_{t-1}}(\x_t\,|\,\z^i_{t-1})~.
\end{equation}

Given the models of motion, death and birth, the approximate prior PHD 
\cite{mahler-zajic01} is estimated from the approximate posterior PHD at the 
previous time instant \cite{mahlerSPIE02} as 
\begin{eqnarray}
\label{eq:phd-prior}
\hat{D}_{\X_t\,|\,\Sigma_{1:t-1}}(\x_t\,|\,Z_{1:t-1}) = \hspace{3.7cm}\nonumber\\
p_B D_{\X_t\,|\,\Sigma_{t-1}}(\x_t\,|\,Z_{t-1}) + \hspace{2.7cm}\nonumber\\
\int (1-p_D)f_{\X_t\,|\,\X_{t-1}}(\x_t\,|\,\x_{t-1})\hspace{1.155cm}\nonumber\\
     \hat{D}_{\X_{t-1}\,|\,\Sigma_{1:t-1}}(\x_{t-1}\,|\,Z_{1:t-1})~d\x_{t-1}~.
\end{eqnarray}

\vspace{-0.1cm}
\paragraph{Observation.} 
We define $p_{FN}$ as the probability that a target 
is {\em not} observed at a given time step (the probability of false 
negative). 
Assuming that there are no spurious observations (a good approximation in
our application where the observations originate from human observers, see 
Section \ref{sec:application-scenario}), the approximate posterior PHD 
distribution is computed \cite{mahlerSPIE02} as
\begin{eqnarray}
\label{eq:phd-posterior}
\hat{D}_{\X_t\,|\,\Sigma_{1:t}}(\x_t\,|\,Z_{1:t}) \approx \hspace{4.3cm}\nonumber\\
\sum_{\z^i_t \in Z_t} 
  f_{\X_t\,|\,\Z_t, \Sigma_{1:t-1}}(\x_t\,|\,\z^i_t, Z_{1:t-1}) +\hspace{1.0cm}\nonumber\\
p_{FN}\hat{D}_{\X_t\,|\,\Sigma_{1:t-1}}(\x_t\,|\,Z_{1:t-1})
\end{eqnarray}
where
\begin{eqnarray}
\label{eq:phd-bayes}
f_{\X_t\,|\,\Z_t, \Sigma_{1:t-1}}(\x_t\,|\,\z^i_t, Z_{1:t-1}) 
\propto \hspace{2.9cm}\nonumber\\
f_{\Z_t\,|\,\X_t}(\z^i_t\,|\,\x_t)~\hat{D}_{\X_t\,|\,\Sigma_{1:t-1}}(\x_t\,|\,Z_{1:t-1})~,
\end{eqnarray}
which is a pdf (with the integral 1 over the state-space).\footnote{Eq
(\ref{eq:phd-posterior}) was wrongly derived in \cite{mahler-zajic01}. 
However, the error was pointed out and corrected in \cite{mahlerSPIE02}.}

Using Eqs (\ref{eq:phd-obs}), (\ref{eq:phd-prior}) and 
(\ref{eq:phd-posterior}), the PHD can be propagated in time. The result of the 
tracking is the estimated number of targets, and the location of the detected 
maxima in the posterior approximate PHD in each time step.

\subsection{Particle implementation}
\label{sec:phd-particle}
 
We will now describe the particle filter implementation of Eqs 
(\ref{eq:phd-obs}), (\ref{eq:phd-prior}) and (\ref{eq:phd-posterior}).
The presentation follows that of the ordinary particle filter (Section
\ref{sec:particle-particle}) to enable comparison.

A pdf (with integral 1) is usually represented with $\mathcal{N}$ particles 
(Section \ref{sec:particle-particle}). 
Here, a PHD (with integral $n_t^X$) is represented with $n_t^X\mathcal{N}$ 
particles, $n_t^X$ being the expected number of targets at time $t$. One 
time-step proceeds as follows:

\vspace{-0.1cm}
\paragraph{Prediction.} 
The posterior PHD at time $t-1$ is represented by a set of (unweighted) 
particles $\{\bxi_{t-1}^1,\ldots ,\bxi_{t-1}^{n^X_{t-1}\mathcal{N}}\}$. These 
are propagated in time by sampling from the dynamical model
$f_{\X_t\,|\,\X_{t-1}}(\x_t\,|\,\bxi^s_{t-1})$ for 
$s = 1,\ldots,n^X_{t-1}\mathcal{N}$. The propagated particles are each given a 
weight $\varpi^s_t = (1-p_D)/\mathcal{N}$. The set of weighted propagated 
particles represent the second term in Eq (\ref{eq:phd-prior}).

Now, for each of the observations 
$\z^i_{t-1}, \in Z_{t-1}, i = 1,\ldots,n^Z_{t-1}$, $\mathcal{N}$ 
particles are sampled from the birth model 
$f_{\X_t\,|\,\Z_{t-1}}(\x_t\,|\,\z^i_{t-1})$ (Eq (\ref{eq:phd-obs})). Each 
particle is
given a weight $\varpi^s_t = p_B/\mathcal{N}$. The resulting set of weighted 
particles represent the first term in Eq (\ref{eq:phd-prior}).

The two weighted particle clouds are concatenated to form a set of particles 
with attached weights, $\{(\tilde{\bxi}_t^1,\varpi_t^1),\ldots ,
(\tilde{\bxi}_t^{(n_{t-1}^Z + n_{t-1}^X)\mathcal{N}},
\varpi_t^{(n_{t-1}^Z + n_{t-1}^X)\mathcal{N}})\}$, that represent the 
approximate prior PHD (Eq (\ref{eq:phd-prior})) at time $t$.

\vspace{-0.1cm}
\paragraph{Observation.} 
For each new observation $\z^i_t \in Z_t, i = 1,\ldots,n_t^Z$, a copy $i$ of 
the prior particle set is made. New weights $\pi_t^{i,s} \propto \varpi^s_t
f_{\Z_t\,|\,\X_t}(\z^i_t\,|\,\tilde{\bxi}_t^s)$ are computed. For each 
set $i$, the
weights are thereafter normalized to sum to one. The re-weighted particle 
set represents the i:th term 
$f_{\X_t\,|\,\Z_t, \Sigma_{1:t-1}}(\x_t\,|\,\z^i_t, Z_{1:t-1})$ in 
the sum in Eq (\ref{eq:phd-posterior}).

The original prior particle set is down-weighted according to  
$\pi^{0,s}_t = p_{FN}\varpi^s_t$. This set now represent the last 
term in Eq (\ref{eq:phd-posterior}).

The concatenation of these sets, $\{(\tilde{\bxi}_t^1,\pi_t^{1}),\ldots,$
$(\tilde{\bxi}_t^{(n_t^Z + 1)(n_{t-1}^Z + n_{t-1}^X)\mathcal{N}},
\pi_t^{(n_t^Z + 1)(n_{t-1}^Z + n_{t-1}^X)\mathcal{N}})\}$, is a weighted
representation of the posterior PHD.

\vspace{-0.1cm}
\paragraph{Resampling.}
An unweighted representation of the posterior PHD is now obtained by 
resampling the weighted particle set.
The expected number of targets is computed as the sum over all weights in 
this set: $n_t^X = \sum_{i = 1}^{(n_t^Z + 1)(n_{t-1}^Z + n_{t-1}^X)}\pi_t^i$.
Now, $n_t^X\mathcal{N}$ new particles are Monte Carlo sampled 
(Section \ref{sec:particle-particle}) from the weighted set. The result is an
unweighted particle set $\{\bxi_t^1,\ldots ,\bxi_t^{n^X_t\mathcal{N}}\}$ that 
represents the approximate posterior PHD 
$\hat{D}_{\X_t\,|\,\Sigma_{1:t}}(\x_t\,|\,Z_{1:t})$ at time $t$.

\section{Terrain application}
\label{sec:application}

The PHD particle filter is here applied to terrain tracking. 
The reason to use particle filtering for terrain tracking is clarified in 
Section \ref{sec:application-motion} -- the motion model of the vehicles is 
non-linear and dependent on the terrain. Using particle filtering, we avoid 
the need to construct an analytical model of the motion noise, since the 
particles provide a sampled representation of the motion distribution.

\subsection{Scenario}
\label{sec:application-scenario}

The scenario is 841 s long, simulated in time-steps of five s. Three 
vehicles (of the same type) travel along roads in the terrain, with 
a normally distributed speed of mean 8.3 m/s and standard deviation 0.1 m/s.
At one time, one of the vehicles travel around 500 m off-road over a field.

The terrain is represented by a discrete map $m$ over position. A pixel in $m$
can take any value $T = \{\mathit{road}, \mathit{field}, \mathit{forest}\}$ 
(exemplified in the tracking movies (Section \ref{sec:experiments}) where 
light grey indicates $\mathit{road}$, white $\mathit{field}$, and grey 
$\mathit{forest}$). 
The probability $p_T(t)$ that a vehicle would select terrain of type $t$ to 
travel in is defined to be $p_T(\mathit{road})=0.66$, 
$p_T(\mathit{field})=0.33$, $p_T(\mathit{forest})=0.01$. 

At each time-step, each vehicle is observed by a human in the terrain with 
probability 0.9, 0.5 or 0.1. 
This means that $p_{FN} = 0.1$ in the first case, $p_{FN} = 0.5$ in the 
second, and $p_{FN} = 0.9$ in the third. For each observation, the observer 
generates a report of the observed vehicle position, 
speed and direction, which is a noisy version of the real state, and of the 
uncertainty with which the observation was made, expressed as standard 
deviation, here $\bsigma_R=[50, 50, 1, \pi/8]$ (m, m, m/s, rad).

\subsection{State-space}
\label{sec:application-space}

The state vector for a vehicle is 
$\x_t = [\p_t, s_t, v_t]$ where $\p_t$ is position (m), $s_t$ speed 
(m/s) and $v_t$ angle (rad). The random set of vehicles is in every time-step 
limited according to $N_t^X \leq 5$ vehicles for computational reasons.

\subsection{Motion model} 
\label{sec:application-motion}

The motion model of the vehicles is
\begin{equation}
\X_t = \X_{t-1} + d\X_{t-1} + \W_t
\end{equation}
where $d\X_{t-1}$ is the movement estimated from the speed and direction in 
$\X_{t-1}$. The noise term is sampled from a distribution which is the product 
of a normal distribution with standard deviation 
$\bsigma_W=[10, 10, 2, \pi/4]$, and of a terrain distribution. The terrain 
distribution depends on probabilities of finding a vehicle in different types 
of terrain. The sampling from this product distribution is implemented as 
follows:
Sample particles $\bxi^i$ using the normally distributed noise term. Each 
particle $i$ now obtains a value $\pi^i = p_T(m(\bxi^i))$. Resample the 
particles according to $\pi^i$ using Monte Carlo sampling.

\subsection{Birth model}
\label{sec:application-birth}

We assume the birth rate $p_B$ and death rate $p_D$ of targets to be invariant 
to position and time-step, and only dependent on the probability of missing 
observations $p_{FN}$. The goal of the tracking is most often to keep track of 
all targets while not significantly overestimating the number of targets. We 
design the birth and death model for this purpose. A high degree of missing 
observations should give a higher birth rate since it 
takes more time steps in general to ``confirm'' a birth with a new 
observation. The mean number of 
steps between observations is $\frac{1}{1 - p_{FN}}$. Therefore, 
\begin{eqnarray}
p_B = K^{1 - p_{FN}} ~,\\
p_D = K~.\hspace{0.98cm}
\end{eqnarray}
The constant $K$ is set empirically to $0.01$.

\subsection{Observation model} 
\label{sec:application-observation}

As mentioned in Section \ref{sec:application-scenario}, observations $\Z_t$ 
are given in the target state-space, which means that 
Eq (\ref{eq:particle-h}) becomes
\begin{equation}
\Z_t = \X_t + \V_t ~.
\end{equation}
The observation noise $\V_t$ is normally distributed with standard deviation
$\bsigma_V = \bsigma_R$ (Section \ref{sec:application-scenario}).

\section{Results}
\label{sec:experiments}

\begin{figure*}[!t]
\vspace{-0.2cm}
\centerline{
\subfigure[$p_{FN} = 0.1$]{\includegraphics[width=5.55cm]{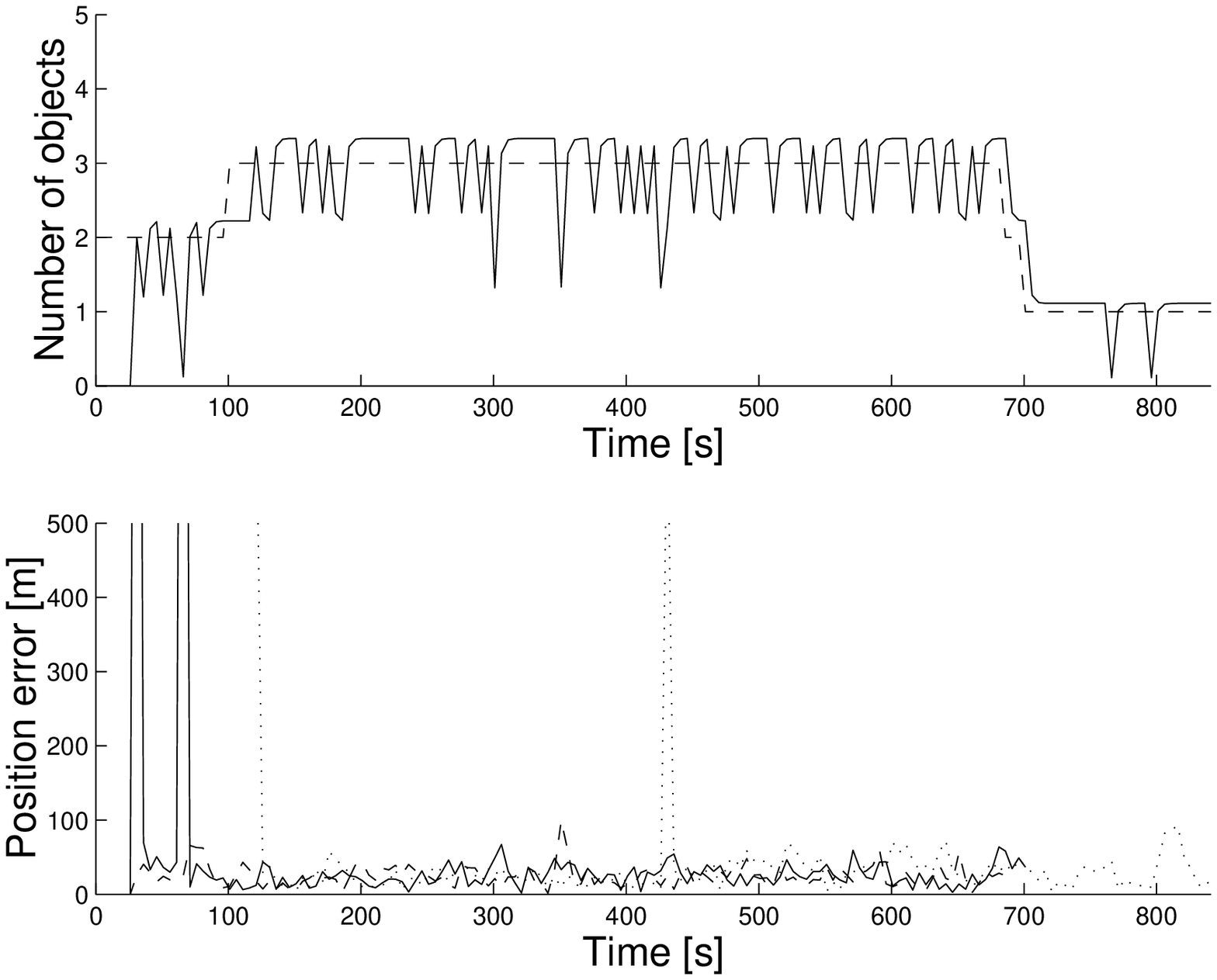}}
\subfigure[$p_{FN} = 0.5$]{\includegraphics[width=5.55cm]{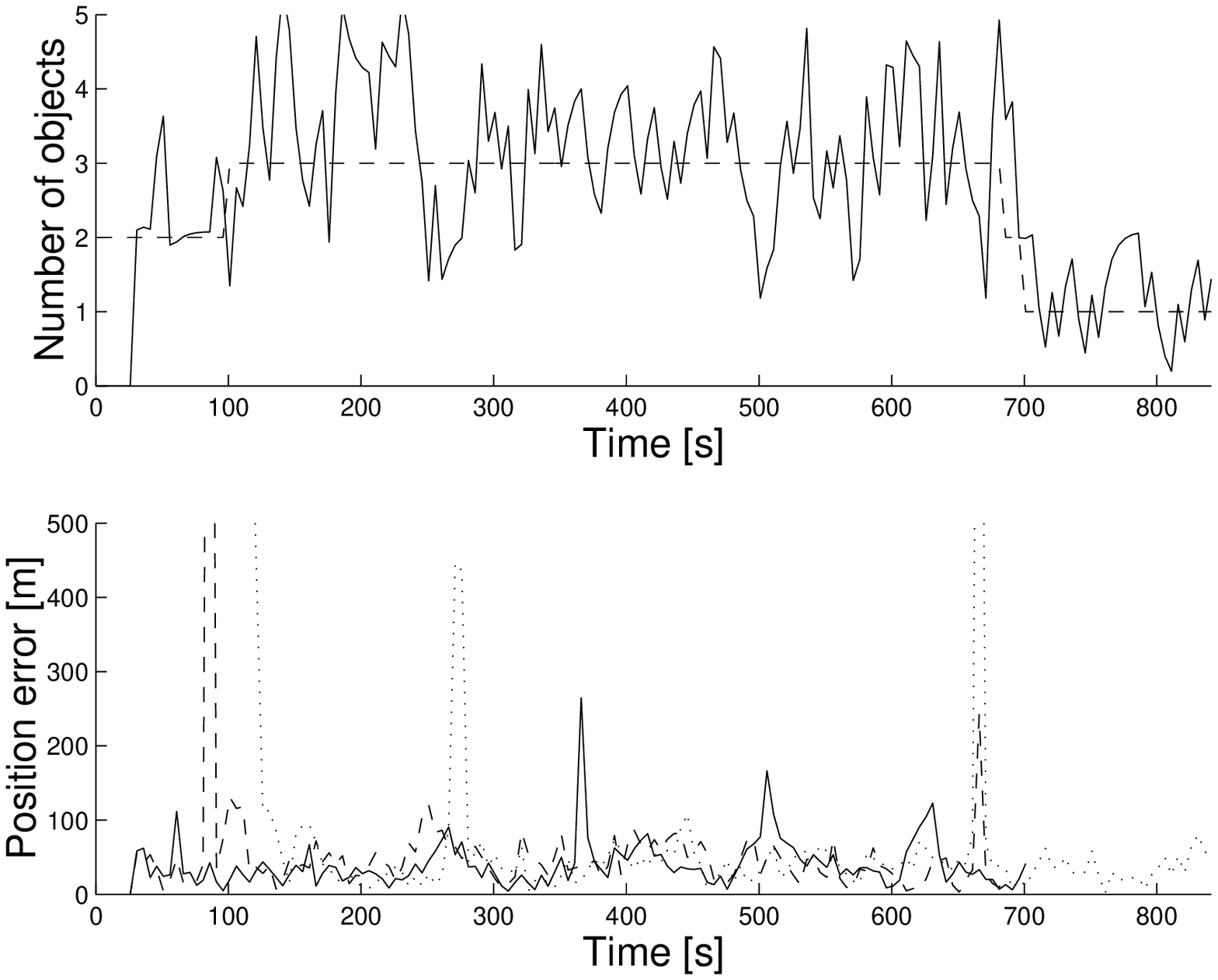}}
\subfigure[$p_{FN} = 0.9$]{\includegraphics[width=5.55cm]{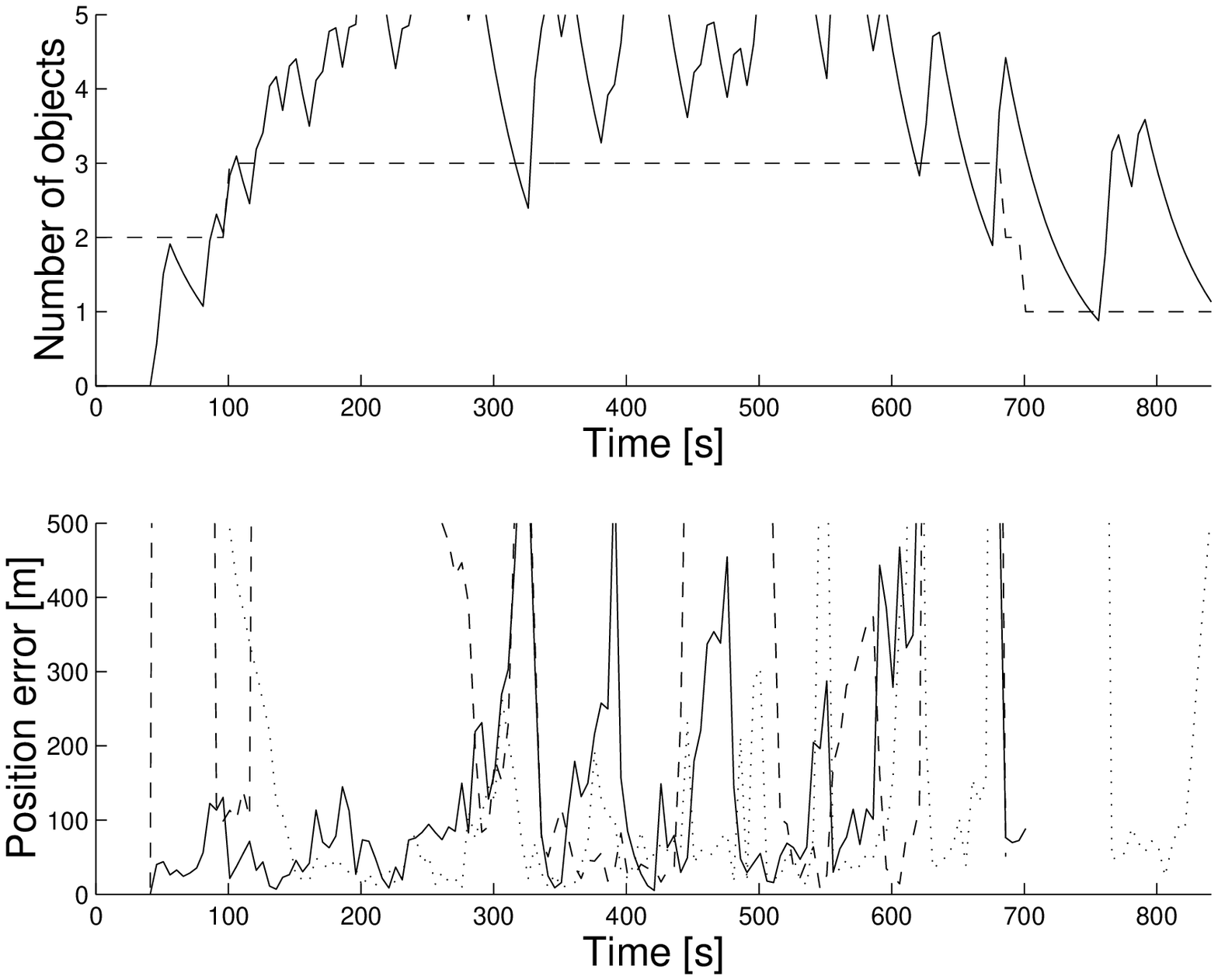}}
}
\caption{\small Tracking errors for the PHD filter. (a) Observation 
probability 0.9. (b) Observation probability 0.5. (c) 
Observation probability 0.1. The upper graph in each subfigure shows 
estimated (solid line) number of targets, compared to the true (dashed line) 
number. The lower graph shows position errors for the three vehicles. Solid, 
dashed and dotted lines denote different vehicles. The dotted target appears 
after 101 s, the dashed target disappears after 687 s and the solid 
target after 702 s. Position error is measured as the Euclidean distance 
from the true target position to the nearest detected maxima in the estimated 
PHD.}
\label{fig:graphsPHD}
\end{figure*}

\begin{figure*}[!t]
\vspace{-0.2cm}
\centerline{
\subfigure[$p_{FN} = 0.1$]{\includegraphics[width=5.55cm]{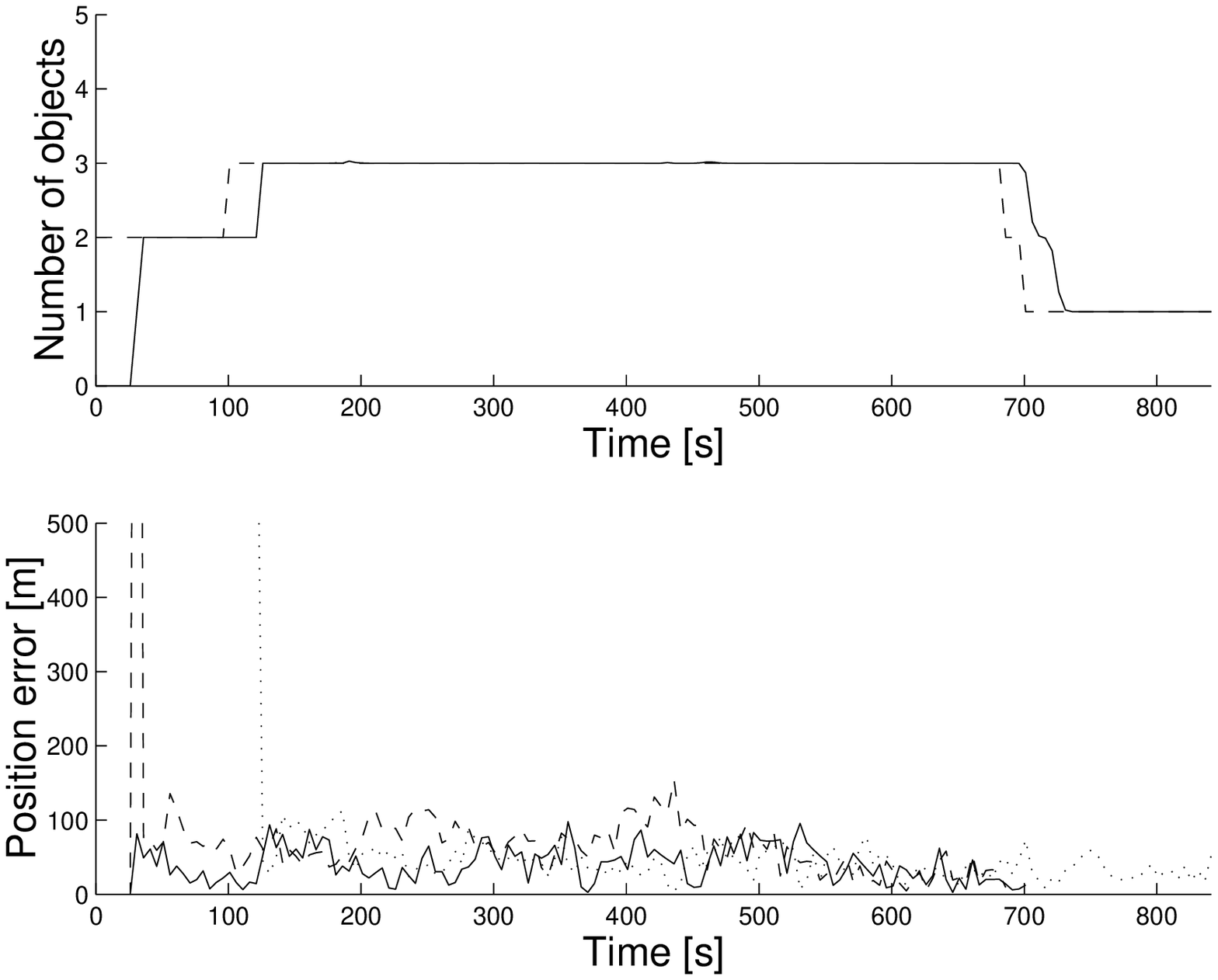}}
\subfigure[$p_{FN} = 0.5$]{\includegraphics[width=5.55cm]{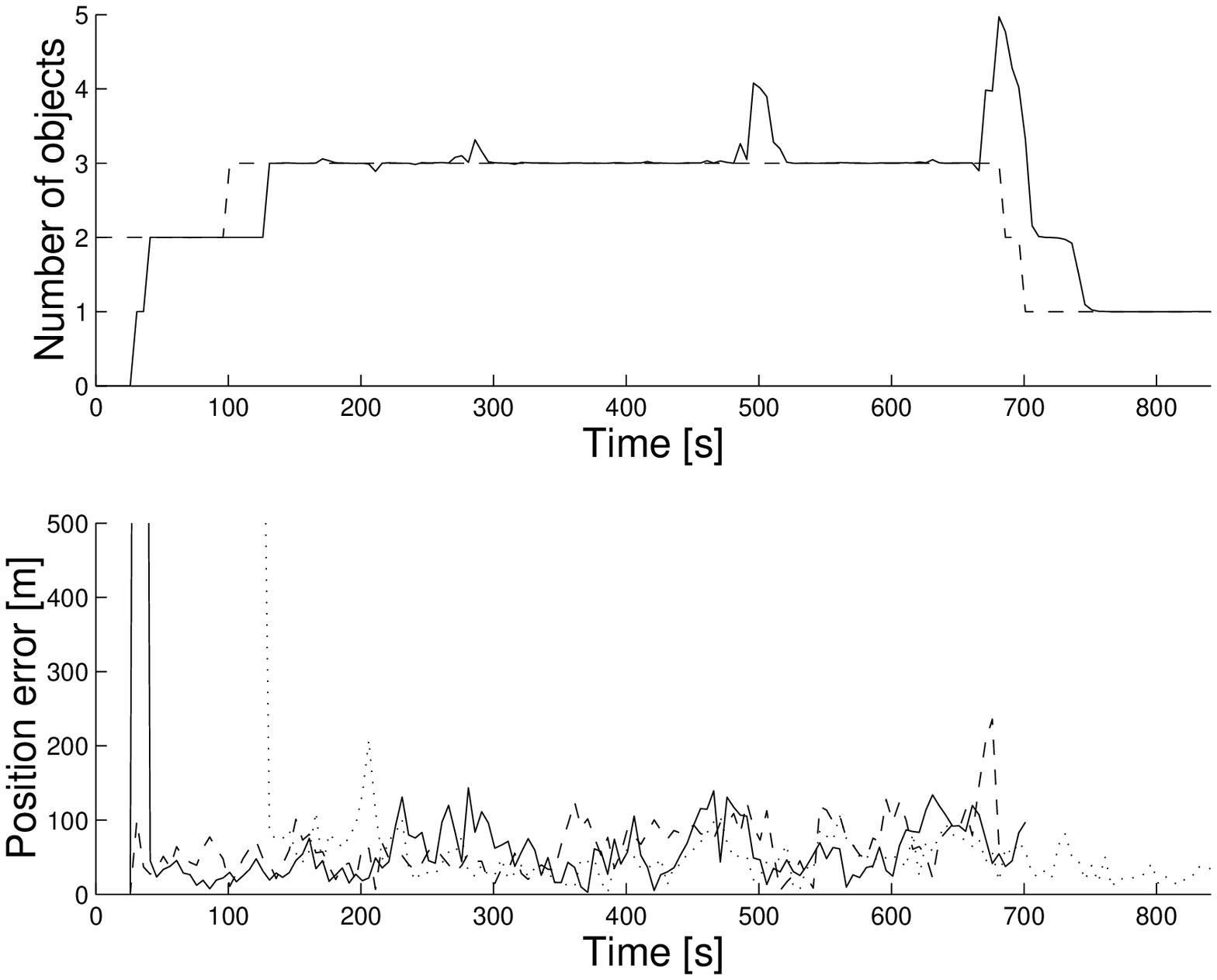}}
\subfigure[$p_{FN} = 0.9$]{\includegraphics[width=5.55cm]{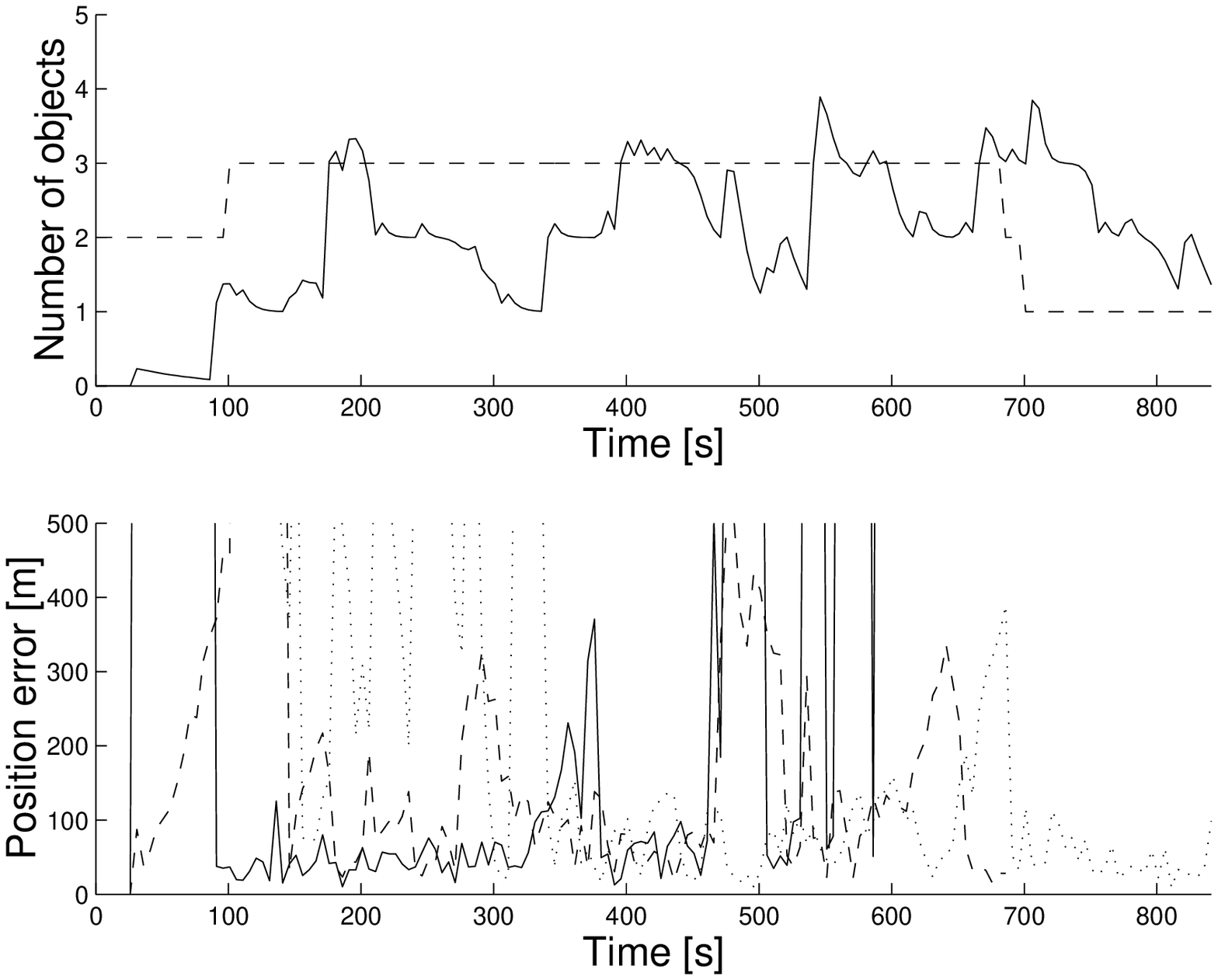}}
}
\caption{\small Tracking errors for the full random set filter, on the same 
scenario. (a) Observation probability 0.9. 
(b) Observation probability 0.5. (c) Observation probability 0.1. A
 complete description of this simulation can be found in 
\cite{sidenbladhWOMOT03}.}
\label{fig:graphsFull}
\end{figure*}

Using the settings described above experiments were performed to test the
performance of the PHD particle filter (Figure \ref{fig:graphsPHD}) and to 
compare it with a particle 
implementation \cite{sidenbladhWOMOT03,sidenbladhTSP03} of the FISST 
filter \cite{mahler-mono}, which maintains the joint distribution over the 
full random set over 
time (Figure \ref{fig:graphsFull}). $\mathcal{N}=1000$ particles were used 
to represent a pdf in the PHD filter. The settings of the FISST particle 
filter simulation can be found in \cite{sidenbladhWOMOT03}.

The tracking performance was measured in two ways, comparing the estimated 
number of targets with the true value, and 
measuring the Euclidean distance between the ground truth target positions 
and the local maxima in the PHD (Section \ref{sec:phd}).
\footnote{Movies of the six tracking examples can be found at 
{\tt\footnotesize http://www.foi.se/fusion/mpg/FUSION03/*.mpg}. 
Two movies relating to each 
of the Figures 1a, 1b, 1c, 2a, 2b, and 2c can be found. For, e.g., Figure 1a, 
the 
movie {\tt\footnotesize phdFigure1(a).mpg} shows the (discretized) PHD
(blue -- 0, red -- 0.2) with white 95\% error ellipses indicating the 
Gaussians fitted to the PHD. 
The movie {\tt\footnotesize terrainFigure1(a).mpg} shows the terrain 
(grey-scale, Section \ref{sec:application-scenario}), the particles (red) and 
Gaussians (deep blue for high PHD peaks, lighter for lower peaks). True 
vehicle positions are indicated by green +, observations by green *.}

Both filters were implemented in Matlab, which is a language not suited for 
real-time applications. However, it should be noted that both algorithms 
required less or marginally more time than the span of a time-step in the 
simulation, 5 s, running in Linux on an ordinary desktop computer. This 
indicates the usability of both algorithms for real-time applications. 

One iteration in the FISST particle filter required 4.9 s on 
average, while an iteration in the PHD particle filter required 0.38 s. 
The generation of the (discretized) PHD and the fitting of the mixture of 
Gaussians to the PHD were identical in the two filters, and required 
1.2 s on average. 
Thus one time-step in the full FISST particle filter takes approximately 12 
times longer than the corresponding iteration in the PHD filter. This should 
be kept in mind while comparing the performance of the two filters. 

As expected, the FISST particle filter outperforms the PHD particle filter in 
estimating the number of targets (upper graph in each subfigure) for all 
tested values of $p_{FN}$. If this is an important aspect of the 
tracking, a filter maintaining belief over the full random set should be 
used.

However, the accuracy in position estimation is very similar between the two 
filters. With high or moderate observation probability (Figures 
\ref{fig:graphsPHD}a,b and \ref{fig:graphsFull}a,b), both filters maintain 
track of all targets, save for a few mistakes in the PHD filter that are 
quickly recovered from. With a low observation probability, both filters 
(Figure \ref{fig:graphsPHD}c and \ref{fig:graphsFull}c) fail 
to track the targets to a high degree. The reasons for that is simply that the 
SNR is too low \cite{mahler-zajic01,sidenbladhWOMOT03}.

To conclude, the PHD particle filter's accuracy in estimating the number 
of targets is low, and falls quickly with the SNR. However, the positions of 
the targets are estimated with the same accuracy as provided by a filter 
representing the full random set.

Thus, the PHD particle filter is a robust and 
computationally inexpensive alternative to representing the full joint 
distribution over the random set,
when estimation of the number of targets is not the primary issue.

\section{Conclusions}
\label{sec:conclusions}

The contribution of this paper has been a particle filtering implementation of 
the PHD filter presented by Mahler and Zajic 
\cite{mahlerSPIE02,mahler-zajic01}. The PHD particle filter was applied to 
tracking of an unknown and changing number of vehicles in terrain, a problem 
incorporating highly non-linear motion, due to the terrain.

Experiments showed the PHD particle filter to be a fast and robust alternative 
to a filter where the full joint distribution over the set of targets was 
maintained over time.

\subsection{Future work}
\label{sec:conclusions-future}

This work could be extended along several avenues of research. Firstly, the 
effects of all parameter settings on the tracking need to be investigated. In 
the experiments in Section \ref{sec:experiments}, only the degree of missing 
observations, $p_{FN}$, was varied.

Furthermore, it would be interesting to investigate more sophisticated 
observation models. The experiments here show clearly 
that the performance of the filter is strongly dependent on the SNR. 
One way to heighten the SNR with our type of sensors, human observers, is to 
take negative information (i.e. absence of reports) into regard. This is 
possible if the fields of view of the observers are known. 

Finally, a real-time implementation should be made, and the filter should be 
tested over longer time periods with more targets. A larger testbed is 
currently developed for this purpose.

\bibliographystyle{plain}
{\small
\bibliography{ref}

\begin{thebibliography}{10}

\bibitem{ballantyne01}
D.~J. Ballantyne, H.~Y. Chan, and M.~A. Kouritzin.
\newblock A branching particle-based nonlinear filter for multi-target
  tracking.
\newblock In {\em International Conference on Information Fusion}, volume~1,
  pages WeA2:3--10, 2001.

\bibitem{doucet01}
A.~Doucet, N.~de~Freitas, and N.~Gordon, editors.
\newblock {\em Sequential Monte Carlo Methods in Practice}.
\newblock Springer Verlag, NEw York, NY, USA, 2001.

\bibitem{fortmann83}
T.~E. Fortmann, Y.~Bar-Shalom, and M.~Scheffe.
\newblock Sonar tracking of multiple targets using joint probabilistic data
  association.
\newblock {\em IEEE Journal of Oceanic Engineering}, OE-8(3):173--184, 1983.

\bibitem{goodman97}
I.~R. Goodman, R.~P.~S. Mahler, and H.~T. Nguyen.
\newblock {\em Mathematics of Data Fusion}.
\newblock Kluwer Academic Publishers, Dordrecht, Netherlands, 1997.

\bibitem{gordon93}
N.~Gordon, D.~Salmond, and A.~Smith.
\newblock A novel approach to nonlinear/non-{Gaussian} {Bayesian} state
  estimation.
\newblock {\em IEE Proceedings on Radar, Sonar and Navigation},
  140(2):107--113, 1993.

\bibitem{hue02}
C.~Hue, J-P.~Le Cadre, and P.~P\'{e}rez.
\newblock Sequential {Monte Carlo} methods for multiple target tracking and
  data fusion.
\newblock {\em IEEE Transactions on Signal Processing}, 50(2):309--325, 2002.

\bibitem{isard-blakeIJCV98}
M.~Isard and A.~Blake.
\newblock Condensation -- conditional density propagation for visual tracking.
\newblock {\em International Journal of Computer Vision}, 29(1):5--28, 1998.

\bibitem{isard-maccormick01}
M.~Isard and J.~MacCormick.
\newblock {BraMBLe}: A {Bayesian} multiple-blob tracker.
\newblock In {\em IEEE International Conference on Computer Vision, ICCV},
  volume~2, pages 34--41, 2001.

\bibitem{kastella00}
K.~Kastella, C.~Kreucher, and M.~A. Pagels.
\newblock Nonlinear filtering for ground target applications.
\newblock In {\em SPIE Conference on Signal and Data Processing of Small
  Targets}, volume 4048, pages 266--276, 2000.

\bibitem{ke00}
C-C. Ke, J.~G. Herrero, and J.~Llinas.
\newblock Comparative analysis of alternative ground target tracking
  techniques.
\newblock In {\em International Conference on Information Fusion}, volume~2,
  pages WeB5:3--10, 2000.

\bibitem{mahler-mono}
R.~Mahler.
\newblock {\em An Introduction to Multisource-Multitarget Statistics and its
  Applications}.
\newblock Lockheed Martin Technical Monograph, 2000.

\bibitem{mahlerSPIE02}
R.~Mahler.
\newblock An extended first-order {Bayes} filter for force aggregation.
\newblock In {\em SPIE Conference on Signal and Data Processing of Small
  Targets}, volume 4729, 2002.

\bibitem{mahler-zajic01}
R.~Mahler and T.~Zajic.
\newblock Multitarget filtering using a multitarget first-order moment
  statistic.
\newblock In {\em SPIE Conference on Signal Processing, Sensor Fusion and
  Target Recognition}, volume 4380, pages 184--195, 2001.

\bibitem{mazor98}
E.~Mazor, A.~Averbuch, Y.~Bar-Shalom, and J.~Dayan.
\newblock {IMM} methods in target tracking: A survey.
\newblock {\em IEEE Transactions on Aerospace and Electronic Systems},
  34(1):103--123, 1998.

\bibitem{musick98}
S.~Musick, K.~Kastella, and R.~Mahler.
\newblock A practical implementation of joint multitarget probabilities.
\newblock In {\em SPIE Conference on Signal Processing, Sensor Fusion and
  Target Recognition}, volume 3374, pages 26--37, 1998.

\bibitem{reid79}
D.~B. Reid.
\newblock An algorithm for tracking mulitple targets.
\newblock {\em IEEE Transactions on Automatic Control}, AC-24(6):843--854,
  1979.

\bibitem{sidenbladhWOMOT03}
H.~Sidenbladh.
\newblock Withheld, currently in double-blind review.
\newblock In {\em IEEE Workshop on Multi-Object Tracking}, 2003, submitted.

\bibitem{sidenbladhTSP03}
H.~Sidenbladh and S-L. Wirkander.
\newblock Particle filtering for random sets.
\newblock {\em IEEE Transactions on Signal Processing}, submitted.

\bibitem{sodtke-llinas01}
E.~P. Sodtke and J.~Llinas.
\newblock Terrain based tracking using position sensors.
\newblock In {\em International Conference on Information Fusion}, volume~2,
  pages ThB1:27--32, 2001.

\bibitem{stone02}
L.~D. Stone.
\newblock A {Bayesian} approach to multiple-target tracking.
\newblock In D.~L. Hall and J.~Llinas, editors, {\em Handbook of Multisensor
  Data Fusion}, 2002.

\end{thebibliography}
}
\end{document}